# Ischemic Stroke Segmentation and Net Water Uptake Quantification on Multicenter Non-Contrast CT Using Supervised Target-Domain Adaptation

Linus Britt[a,b] , Maximilian Nielsen MSc[a,b] , Susan Klapproth MD[c], André Kemmling MD[d], Michael H. Lev MD[e], Gabriel Broocks MD[f,g], René Werner PhD[a,b], Thilo Sentker PhD[a,b,1]

[a]*Institute for Applied Medical Informatics, University Medical Center Hamburg-Eppendorf, Hamburg, Germany*
[b]*Institute of Computational Neuroscience, University Medical Center Hamburg-Eppendorf, Hamburg, Germany*
[c]*Department of Diagnostic and Interventional Neuroradiology, University Medical Center Hamburg-Eppendorf, Hamburg, Germany*
[d]Department of Neuroradiology, University Hospital Marburg, Marburg, Germany
[e]*Department of Radiology, Massachusetts General Hospital, Harvard Medical School, Boston, MA, United States of America*
[f]*Department of Neuroradiology, HELIOS Medical Center Schwerin, University Campus of MSH Medical School Hamburg, Schwerin, Germany*
[g]*MSH Research, Development and Innovation GmbH, MSH Medical University of Applied Sciences and Medical University, Hamburg, Germany*

---

[1] Corresponding author
*Address:* Martinistr. 52, 20246 Hamburg, Germany
*Phone:* +49 (0) 40 7410-58026
*Fax:* +49 (0) 40 7410-54882

*Email address:* t.sentker@uke.de (Thilo Sentker)

**Abstract**

**Objectives:**

Quantitative assessment of infarct hypodensity on non-contrast computed tomography (NCCT), including net water uptake (NWU), requires manual or semi-manual lesion delineation, often guided by CT perfusion or diffusion-weighted MRI, limiting clinical applicability. Automated NCCT segmentation could enable such biomarker extraction but remains challenging across heterogeneous multicenter data. This study developed and externally tested a deep learning framework for ischemic stroke segmentation and NWU quantification on NCCT.

**Materials & Methods:**

In this retrospective multicenter study of 801 patients from four datasets, an nnU-Net-based model with foundation-model initialization, extended by modules comparing local features with the contralateral hemisphere, was trained on NCCT scans from the University Medical Center Hamburg-Eppendorf and the Acute Ischemic Stroke Dataset. The model was fine-tuned on target-domain subsets of the Boston (n=11) and ISLES (n=75) datasets, with evaluation on held-out cases. Segmentations and NWU were compared with expert references.

**Results:**

Internally, mean Dice increased from 0.53 (conventional nnU-Net) to 0.57 (proposed approach). After fine-tuning, the proposed approach achieved median Dice of 0.56 for acute lesion segmentation (Boston) and 0.26 for post-treatment final-infarct prediction (ISLES), where small lesions predominated. For lesions ≥30 mL, median Dice increased to 0.67 and 0.57, respectively. On the Boston dataset, automated and reference NWU agreed (concordance correlation coefficient 0.93, mean absolute error 1.24 percentage points).

**Conclusion:**

Contralateral feature comparison and target-domain adaptation supported NCCT-only infarct segmentation across heterogeneous external cohorts, although performance was lower for post-treatment final-infarct prediction. The approach enabled low-error NWU quantification from baseline NCCT without advanced imaging or manual annotation.

**Key Points:**

*Question:* Automated stroke segmentation on non-contrast CT remains challenging due to subtle infarct visibility and multicenter acquisition variability, limiting scalable quantification of segmentation-derived clinical imaging biomarkers.

*Findings:* Embedding hemispheric comparison as medical domain knowledge into deep learning enabled infarct segmentation with endpoint-dependent accuracy and low-error net water uptake quantification from non-contrast CT.

*Critical Relevance Statement:* The framework enables automated stroke segmentation and net water uptake assessment using baseline imaging and adapts to new sites with small annotated subsets. By providing quantitative information without advanced imaging, it creates a basis for accessible prognostic stroke evaluation.

**List of Abbreviations**

**ADC:** Apparent diffusion coefficient

**AIS:** Acute ischemic stroke

**ASPECTS:** Alberta Stroke Program Early CT Score

**CBV:** Cerebral blood volume

**CCC:** Concordance correlation coefficient

**CI:** Confidence interval

**CTP:** Computed tomography perfusion

**DL:** Deep learning

**DWI:** Diffusion-weighted imaging

**HD95:** Hausdorff distance ($95^{th}$ percentile)

**HU:** Hounsfield unit

**IoU:** Intersection over union

**IQR:** Interquartile range

**LDR:** Lesion detection rate

**MAE:** Mean absolute error

**mAP:** Mean average precision

**NCCT:** Non-contrast computed tomography

**NWU:** Net water uptake

**SD:** Standard deviation

**TTD:** Time-to-drain

## 1. Introduction

Rapid treatment selection is critical in acute ischemic stroke (AIS), where timely reperfusion strongly influences outcome [1]. Non-contrast computed tomography (NCCT) is the first-line imaging modality because it is rapid, widely available, and excludes intracranial hemorrhage. It supports patient selection for intravenous thrombolysis and endovascular thrombectomy [2]. NCCT has also gained importance in late and follow-up windows, when computed tomography perfusion (CTP) may underestimate ischemic tissue and NCCT hypodensity can complement infarct core assessment [3].

However, NCCT-based ischemic tissue segmentation remains challenging, as early infarct changes are subtle, with only minor intensity differences relative to normal tissue [4]. Reliable lesion delineation therefore depends on manual or semi-manual annotation, guided by CTP or diffusion-weighted imaging (DWI), which is time-consuming, prone to inter-rater variability, and limits the use of prognostic NCCT-derived biomarkers such as net water uptake (NWU) [5–7]. Robust automated segmentation is therefore needed for broader clinical translation of NWU.

NWU reflects ischemic tissue water accumulation and has been associated with malignant edema, lesion progression, and functional outcome [6, 7]. Recent work further linked NWU to treatment effect in low-ASPECTS stroke [8]. Although NWU has previously been approximated using classical image analysis [9], moderate segmentation performance motivates deep learning (DL)-based approaches – not only for acute infarct delineation and NWU quantification, but also for predicting post-treatment infarct extent from baseline NCCT to support early prognostication.

However, limited generalizability remains a major challenge for DL models [10]. NCCT images vary across institutions due to differing scanners, acquisition protocols, and reconstruction settings. The resulting domain shift can degrade segmentation performance at unseen sites [11, 12].

Mitigating this vulnerability requires explicit medical domain knowledge, such as anatomical priors invariant across acquisition settings. Radiologists naturally rely on this principle when assessing subtle hypodensity against the contralateral hemisphere, an intra-scan reference inherently robust to domain shift. While supplying networks with a mirrored image copy exploits this reference (a strategy that secured us second place in the multicenter ISLES 2026 challenge on T1-weighted MRI[2]), mirrored input alone neither establishes local correspondence nor reliably separates pathology from physiological asymmetry or misalignment.

To overcome the limitations of naïve mirroring, we integrate explicit local feature comparison into a fully automated nnU-Net-based [13] pipeline for ischemic stroke segmentation on NCCT. Our approach combines radiological foundation-model initialization with modules that contrast native and mirrored representations, followed by selective fine-tuning on small target-domain subsets. We evaluate this framework on acute lesion segmentation, downstream NWU quantification, and post-treatment infarct prediction from preinterventional NCCT.

[2] Official challenge result; the publication reporting the challenge results is pending.

## 2. Materials and Methods

### 2.1 Datasets

This retrospective study was approved by the institutional review boards of the University Medical Center Hamburg-Eppendorf (UKE) and Massachusetts General Hospital, Boston. Informed consent was waived because the data were anonymized, and public datasets were used under their respective policies. Reporting followed the 2024 Checklist for Artificial Intelligence in Medical Imaging (Supplementary Material). Cohort characteristics and reference standards are summarized in Table 1.

The UKE dataset and the Acute Ischemic Stroke Dataset (AISD) were used for model development and five-fold internal testing, whereas the Boston and ISLES 2024 datasets were each divided into separate fine-tuning and held-out test sets (Fig. 1).

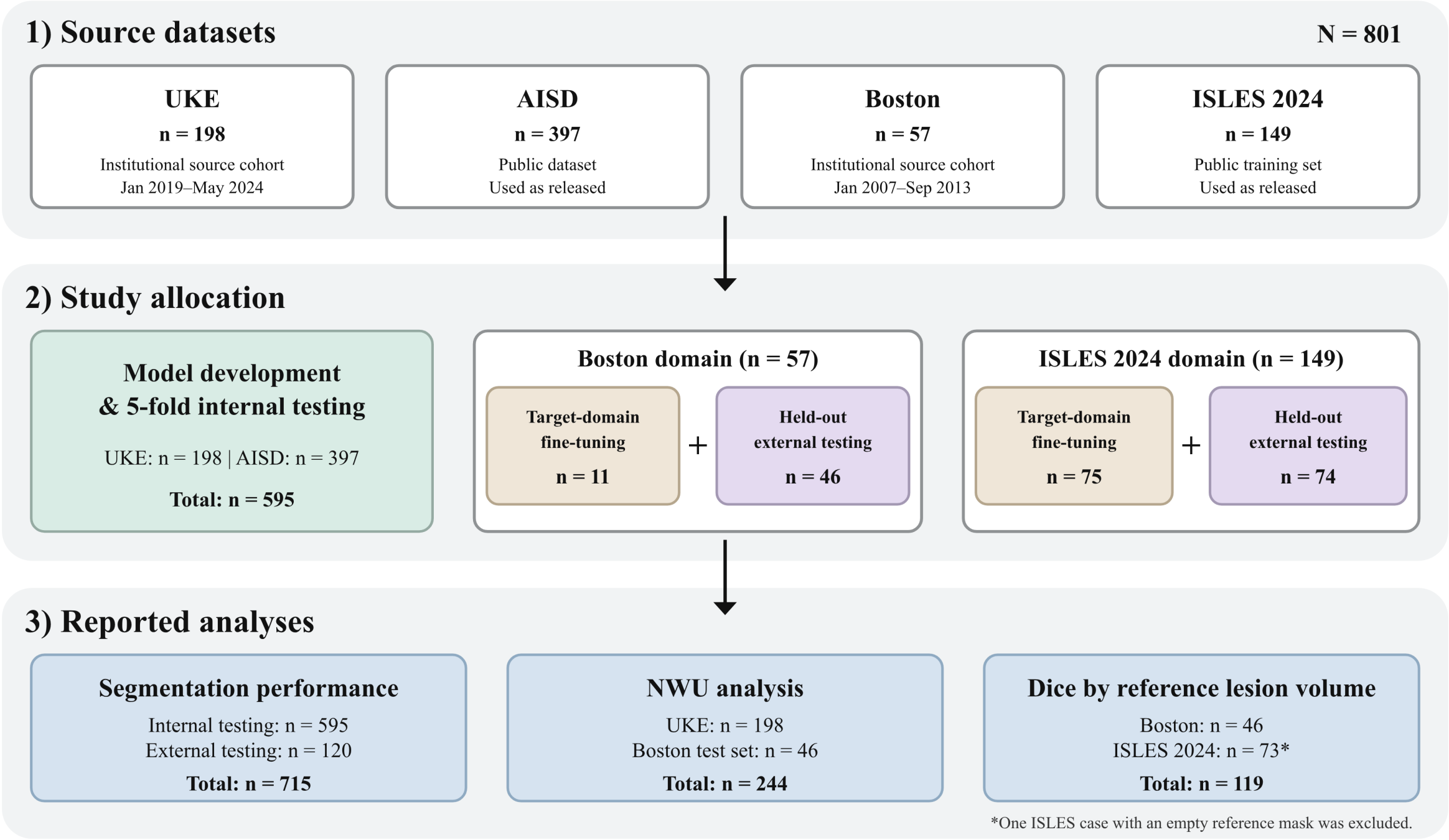

**Figure 1:** Study cohort composition, allocation, and analysis populations. Colors distinguish study roles. One ISLES 2024 case with an empty reference mask was excluded from the lesion-volume-stratified Dice analysis. AISD, Acute Ischemic Stroke Dataset; ISLES, Ischemic Stroke Lesion Segmentation; NWU, net water uptake; UKE, University Medical Center Hamburg-Eppendorf.

- **UKE Dataset:** This dataset comprises 198 patients with AIS in the middle cerebral artery territory, imaged within 6 hours of symptom onset. Although the proposed framework segments exclusively on NCCT, reference lesions were defined by an expert with over five years of experience using pretreatment CTP-derived time-to-drain (TTD) and cerebral blood volume (CBV) maps.
- **AISD [14]:** A public dataset featuring 397 NCCT scans acquired within 24 hours of onset, with reference masks annotated using DWI acquired within the following 24 hours.
- **Boston Dataset:** The dataset included 57 eligible cases with baseline NCCT acquired within 6 hours of onset. Reference masks were derived by clinical experts with more than five years of experience from DWI and apparent diffusion coefficient (ADC) maps acquired after NCCT but before the treatment decision.
- **ISLES 2024 Dataset [15, 16]:** The public training set comprised 149 cases with preinterventional NCCT (<24 h after onset) followed by successful intracranial reperfusion. Final-infarct masks were derived from DWI/ADC 2–9 days after intervention and affinely registered to NCCT space.

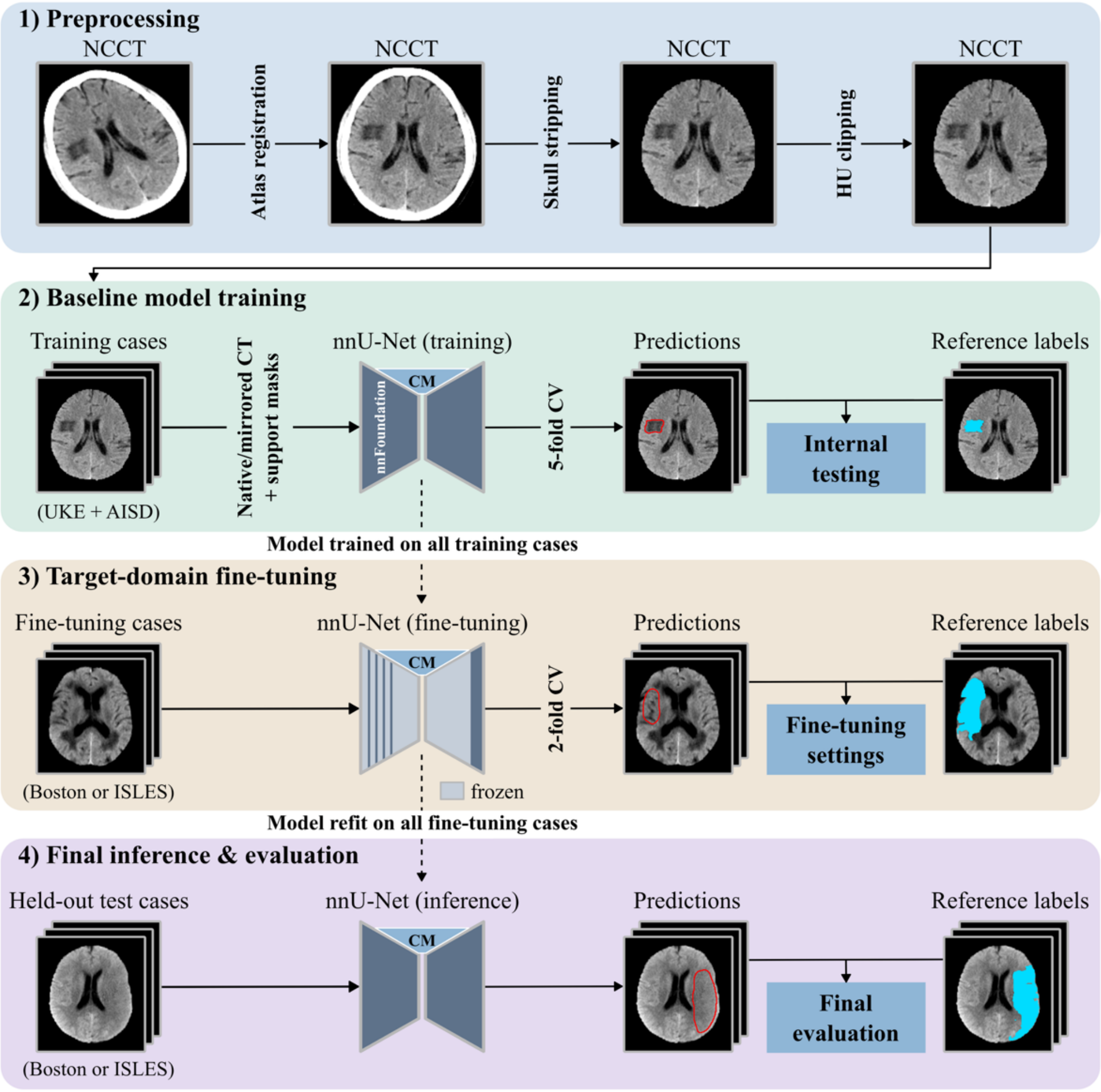

**Figure 2:** Overview of the proposed infarct segmentation workflow. 1) NCCT preprocessing with atlas registration, skull stripping, and HU clipping. 2) Native and mirrored NCCT with support masks are processed by an nnU-Net with a shared, nnFoundation-initialized encoder and comparison modules (CM). Five-fold CV yields internal test predictions; a model trained on all training cases serves as the source for fine-tuning. 3) For each target domain, only the CM, the last decoder block, and early encoder normalization parameters are fine-tuned (lighter components frozen). Fine-tuning settings are selected by two-fold CV, followed by refitting on all fine-tuning cases. 4) The fine-tuned model is applied to the held-out test set and evaluated against reference labels. CM, comparison module; CV, cross-validation; HU, Hounsfield unit; NCCT, non-contrast CT; UKE, University Medical Center Hamburg-Eppendorf; AISD, Acute Ischemic Stroke Dataset.

### 2.2 Preprocessing

Within the overall workflow (Fig. 2), all NCCT scans underwent standardized preprocessing before training and inference. First, an affine registration to a common atlas [17] aligned the images across patients and datasets, with linear resampling to a uniform voxel spacing of 1.0 × 1.0 × 3.0 $mm^3$. TotalSegmentator [18] was then used for skull stripping to restrict analysis to brain tissue. Finally, Hounsfield unit (HU) values were clipped to [0, 80] to suppress extreme intensities and enhance contrast between ischemic lesions and surrounding tissue.

### 2.3 Model Architecture and Training

Models were implemented in the nnU-Net framework (version 2.8.1) [13]. Training and inference used the 3d_fullres configuration and the medium-sized Residual Encoder U-Net preset, with the encoder initialized from the pretrained radiological foundation model nnFoundation [19]. Native NCCT and a left–right mirrored copy were processed by this encoder with shared weights to exploit hemispheric symmetry.

Inspired by local contralateral feature matching [20], two comparison modules at early, high-resolution encoder stages compared each native feature with similarity-weighted mirrored features from a 3 × 3 in-plane neighborhood, tolerating minor residual misalignment. They combined feature disparities with local intensity differences and refined the corresponding decoder skip connections through learned, gated residual corrections. Binary support masks of both images (HU > 0 after clipping), requiring no annotation, excluded comparisons outside valid image regions.

Baseline models were trained on the combined UKE and AISD data using five-fold cross-validation with patient-level splitting, yielding five independent networks whose out-of-fold predictions were used for internal testing. An additional model trained on all development cases served as the source for target-domain fine-tuning. Training used AdamW, a combined Dice and cross-entropy loss, and 100 epochs, with learning rates of $1 \times 10^{-4}$ for the pretrained encoder and $1 \times 10^{-3}$ for the remaining network. Data augmentation was identical for both views; all other parameters followed the default nnU-Net configuration (Fig. 2).

### 2.4 Target-Domain Fine-Tuning

Following baseline training, the all-cases model was separately fine-tuned for each external dataset, with adaptation and hyperparameter selection confined to the corresponding fine-tuning set.

To preserve comparability with [9], we retained the same 46 NCCT cases as the Boston test set and used the remaining 11 for fine-tuning. For ISLES, a pragmatic approximately equal random split yielded 75 fine-tuning and 74 test cases.

Fine-tuning was restricted to the comparison modules and the final decoder block with its segmentation head, together with affine normalization parameters of the early encoder stages to accommodate acquisition-related appearance shifts [21, 22]. The remaining encoder and decoder weights were frozen to preserve the learned feature representations and prevent catastrophic forgetting. Fine-tuning used 10 and 40 epochs for the Boston and ISLES datasets, respectively, with 20 iterations per epoch and AdamW.

We used constant learning rates of $1 \times 10^{-3}$ for the comparison modules and decoder components and $5 \times 10^{-4}$ for the normalization parameters. Learning rates and fine-tuning duration were determined by two-fold cross-validation within the fine-tuning sets, followed by refitting on the complete fine-tuning set (Fig. 2).

### 2.5 NWU Computation

Predicted and reference masks were mapped to native NCCT space, and NWU was computed from the original, unclipped HU images. Mean HU values were extracted from the lesion and the contralateral region obtained by mirroring the mask across the midsagittal plane after interhemispheric registration. NWU was calculated as:

$$NWU = \left(1 - \frac{\overline{HU}_{\text{ischemic}}}{\overline{HU}_{\text{contralateral}}}\right) \times 100$$

where $\overline{HU}_{\text{ischemic}}$ and $\overline{HU}_{\text{contralateral}}$ are the average HU values of the ischemic region and the contralateral reference region, respectively. NWU was evaluated only in cohorts with admission NCCT within 6 h of known symptom onset (UKE, Boston), consistent with its use as an early admission edema biomarker [6]. AISD (NCCT up to 24 h) and ISLES (onset-to-imaging time not derivable in 79/149 cases [53%] and postinterventional final-infarct references) were excluded to avoid heterogeneous imaging windows and, for ISLES, an unsuitable reference endpoint for baseline edema assessment.

NWU was calculated whenever the predicted lesion mask was non-empty, regardless of segmentation overlap. All predicted UKE and Boston lesion maps were non-empty and therefore included.

**2.6 Experiments and Inference Strategy**

For final inference, each fine-tuned model was applied with nnU-Net test-time augmentation. A standard probability threshold of 0.5 was then applied to obtain binary masks for segmentation and NWU evaluation (Fig. 2).

To assess the effects of individual pipeline components, the native single-channel nnU-Net served as a conventional DL baseline and was compared on identical internal folds with configurations stepwise adding mirrored input, nnFoundation initialization, comparison modules, and support masks. Without comparison modules, mirrored images were concatenated as an input channel, and each initialization used its own schedule (250 epochs from scratch; 100 with nnFoundation). All configurations were also evaluated on the Boston dataset before and after fine-tuning.

**2.7 Evaluation Metrics and Statistical Analysis**

Segmentation performance was evaluated using the Dice similarity coefficient, lesion detection rate (LDR), mean average precision (mAP), and 95th percentile Hausdorff distance (HD95). LDR was defined as any overlap between predicted and reference lesions (≥1 voxel) [9], reflecting lesion localization rather than delineation accuracy. mAP was computed as the proportion of cases exceeding an intersection-over-union (IoU) threshold of $0.5^{3/2}$ (approximately 0.35) for comparability with [9], using the metric framework in [23]. HD95 was calculated for all cases with non-empty predictions and reference masks, irrespective of overlap. In the UKE and Boston datasets, NWU error was quantified using mean absolute error (MAE), and agreement using Lin's concordance correlation coefficient (CCC) with 95% confidence intervals (CIs) and Bland–Altman analysis. Bland–Altman differences were defined as automated minus reference NWU and reported as mean bias and 95% limits of agreement (bias ± 1.96 × standard deviation (SD)) in percentage points. Dice was summarized using the same reference lesion-volume strata in both external test cohorts. Associations of lesion volume with Dice (ISLES dataset) and absolute NWU error (Boston dataset) were assessed using Spearman's rank correlation. All CIs were estimated using percentile bootstrapping with 10,000 paired case-level resamples. Cases with empty reference masks were excluded from lesion-volume analyses. Analyses used Python 3.10.18, PyTorch 2.8.0, NumPy 2.2.6, SciPy 1.15.3, and SimpleITK 2.5.2.

## 3. Results

Of 801 patients, 595 were used for model development/internal testing, 86 for fine-tuning, and 120 for held-out external testing (Fig. 1). Table 1 summarizes clinical and imaging heterogeneity across cohorts.

**Table 1:** Clinical and demographic characteristics of patient cohorts.

| Patient characteristics | UKE | AISD | Boston | ISLES 2024 |
| --- | --- | --- | --- | --- |
| Number of patients | n = 198 | n = 397 | n = 57 | n = 149 |
| Age (median [IQR]) | 76 [64–83] | N/A[b] | 68 [54–78] | 76 [62–82] |
| Female (n [%]) | 94 [47] | 129 [33] | 24 [42] | 73 [49] |
| NIHSS on admission (median [IQR]) | 15 [9–19] | N/A[b] | 11 [8–19] | 10 [6–16] |
| Symptom onset – admission in h (median [IQR]) | 3.1 [1.5–5.9] | N/A[b] | 3.5 [2.2–5.3] | 1.7 [1.1–2.6][a] |
| ASPECTS (median [IQR]) | 8 [6–9] | N/A[b] | 8 [7–9] | N/A[b] |
| NCCT in-plane spacing in mm (range) | 1.00 | 0.36–0.58 | 0.43 | 0.30–0.56 |
| NCCT slice spacing in mm (range) | 1.00 | 2.60–10.08 | 5.00 | 0.80–4.00 |
| Reference imaging | CTP (TTD/CBV) | DWI | DWI/ADC | DWI |
| Reference timing | Pretreatment | ≤24h after NCCT | Pretreatment | 2-9 days postintervention |
| Mask lesion volume in mL (median [IQR]) | 67.3 [52.1–81.7] | 16.3 [5.7–40.7] | 31.4 [9.1–74.0] | 12.4 [3.9–38.6] |
| Lesions < 10mL (n [%]) | 1 [0.5] | 146 [37] | 16 [28] | 66 [44] |
| Multi-lesion cases (n [%]) | 7 [4] | 295 [74] | 1 [2] | 131 [88] |

[a] Symptom onset to admission time was available for 74/149 patients in the ISLES cohort.

[b] N/A, variable not available in the publicly released dataset.

### 3.1 Internal Testing and Boston Acute Lesion Segmentation

Table 2 summarizes internal and external segmentation performance. During internal cross-validation, the baseline model achieved a median Dice score of 0.64 (interquartile range (IQR): 0.41–0.78). To compare with Liang et al. [14], we evaluated their AISD test subset using only cross-validation predictions. Mean Dice was 0.66, exceeding the 0.58 reported in the original study.

For acute lesion segmentation after adaptation to the external Boston domain, the fine-tuned model yielded a median Dice score of 0.56 (IQR 0.44–0.70), compared with 0.45 before fine-tuning. LDR remained high during internal evaluation (94.3%) and on the external Boston test set (95.7%). Mean HD95 values were 27.38 mm internally and 18.34 mm externally, with corresponding mAP values of 0.66 and 0.63.

### 3.2 ISLES Lesion Segmentation and Comparison to Post-treatment Infarction

For the second use case, the ISLES-adapted model predicted DWI-defined post-treatment infarct extent from preinterventional NCCT in the held-out subset, in line with the ISLES 2024 challenge goal.

The ISLES-adapted model achieved a median Dice of 0.26 (IQR 0.01–0.45) and a mean of 0.27 (SD 0.26). For detection, LDR was 77.0% and mAP 0.20. After excluding one case with an empty final-infarct mask, final-infarct volume was strongly associated with Dice (n=73; Spearman's $\rho = 0.75$, 95% CI 0.60–0.85; $p < 0.001$). For lesions ≥30 mL (Table 3), median Dice was 0.57 (IQR 0.43–0.70, n=20), compared with 0.67 (IQR 0.54–0.74, n=26) for acute lesion segmentation on the Boston dataset.

**Table 2:** Model performance across internal and external cohorts. Results are shown for the combined internal testing (UKE + AISD) and the two external testing (Boston and ISLES) datasets. Segmentation performance is summarized using lesion detection rate (LDR, %), mean average precision (mAP), Dice similarity coefficient (median [IQR] and mean (SD)), and 95th percentile Hausdorff distance (HD95 [mm]; mean).

| Stage | Dataset | n | LDR (%) (Dice > 0) | mAP (IoU > 0.5$^{3/2}$) | Dice (median [IQR]) | Dice (mean (SD)) | HD95 [mm] (mean) |
|---|---|---|---|---|---|---|---|
| Internal testing | UKE + AISD | 595 | 94.3 | 0.66 | 0.64 [0.41–0.78] | 0.57 (0.27) | 27.38 |
| External testing | Boston | 46 | 95.7 | 0.63 | 0.56 [0.44–0.70] | 0.53 (0.22) | 18.34 |
| External testing | ISLES | 74 | 77.0 | 0.20 | 0.26 [0.01–0.45] | 0.27 (0.26) | 43.56 |

**Table 3:** Dice stratified by reference lesion volume for Boston lesion segmentation and ISLES final-infarct prediction. Values are median Dice similarity coefficient (IQR) and number of cases per stratum. One ISLES case with an empty reference mask was excluded.

| Reference lesion volume | Boston (n=46) | | ISLES 2024 (n=73) | |
|---|---|---|---|---|
| | n | Dice (median [IQR]) | n | Dice (median [IQR]) |
| <5 mL | 5 | 0.04 [0.00–0.20] | 25 | 0.03 [0.00–0.04] |
| 5 – <15 mL | 9 | 0.51 [0.44–0.55] | 18 | 0.26 [0.01–0.37] |
| 15 – <30 mL | 6 | 0.48 [0.37–0.60] | 10 | 0.44 [0.34–0.55] |
| 30 – <70 mL | 12 | 0.56 [0.51–0.64] | 8 | 0.50 [0.21–0.62] |
| ≥70 mL | 14 | 0.72 [0.66–0.80] | 12 | 0.68 [0.45–0.71] |

Figures 3 and 4 show internal and external examples across the Dice range. Lower scores mainly occurred with small or multifocal lesions.

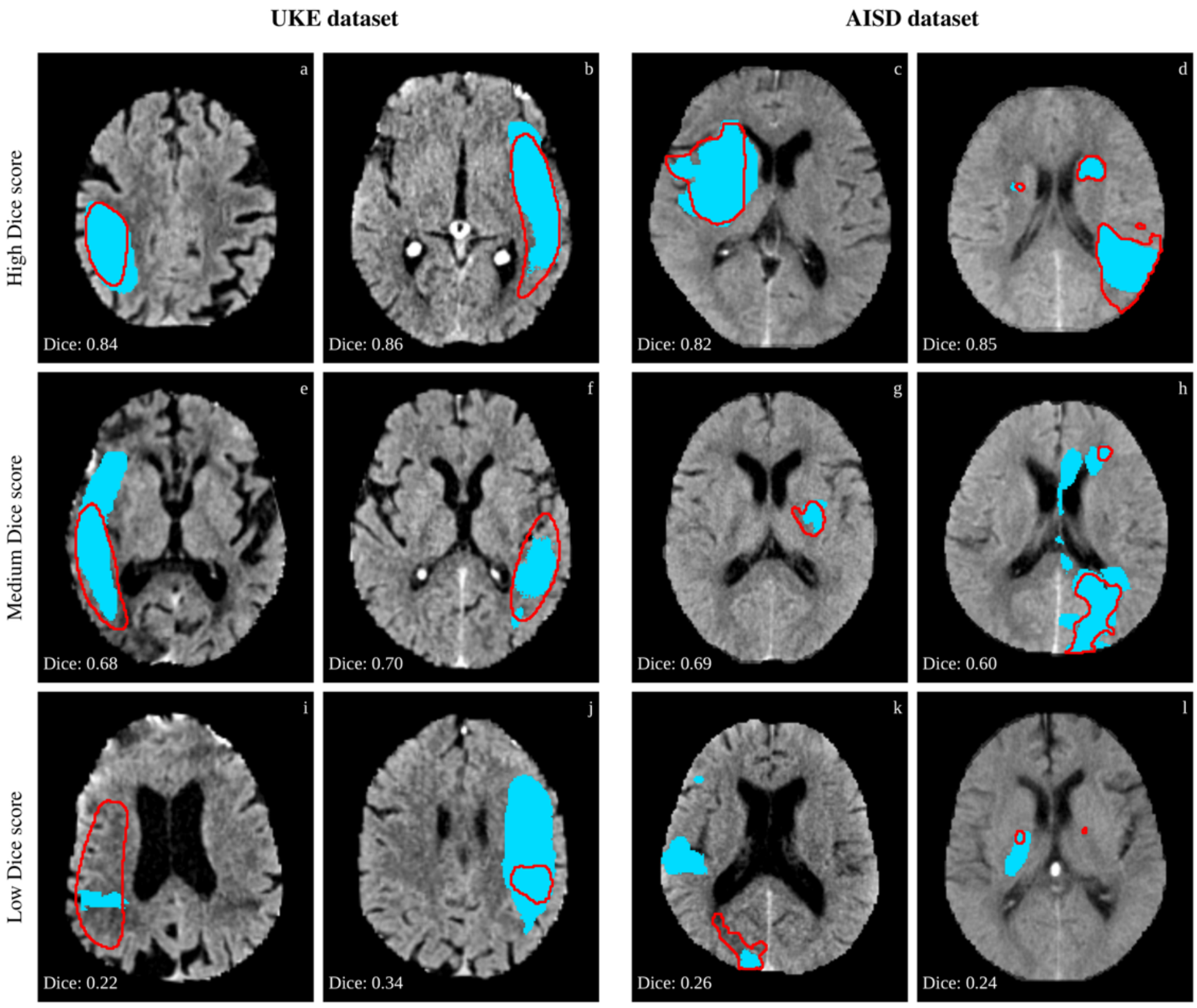

**Figure 3:** Twelve representative cases from the internal testing data are shown, including six from the UKE dataset and six from the AISD dataset, covering a range of segmentation performance levels (high, medium, and low Dice scores). For each case, the NCCT image is displayed with the reference lesion mask shown in cyan and the predicted segmentation outlined in red. Dice similarity coefficients are reported in the bottom left corner, and case identifiers are shown in the top right corner of each image.

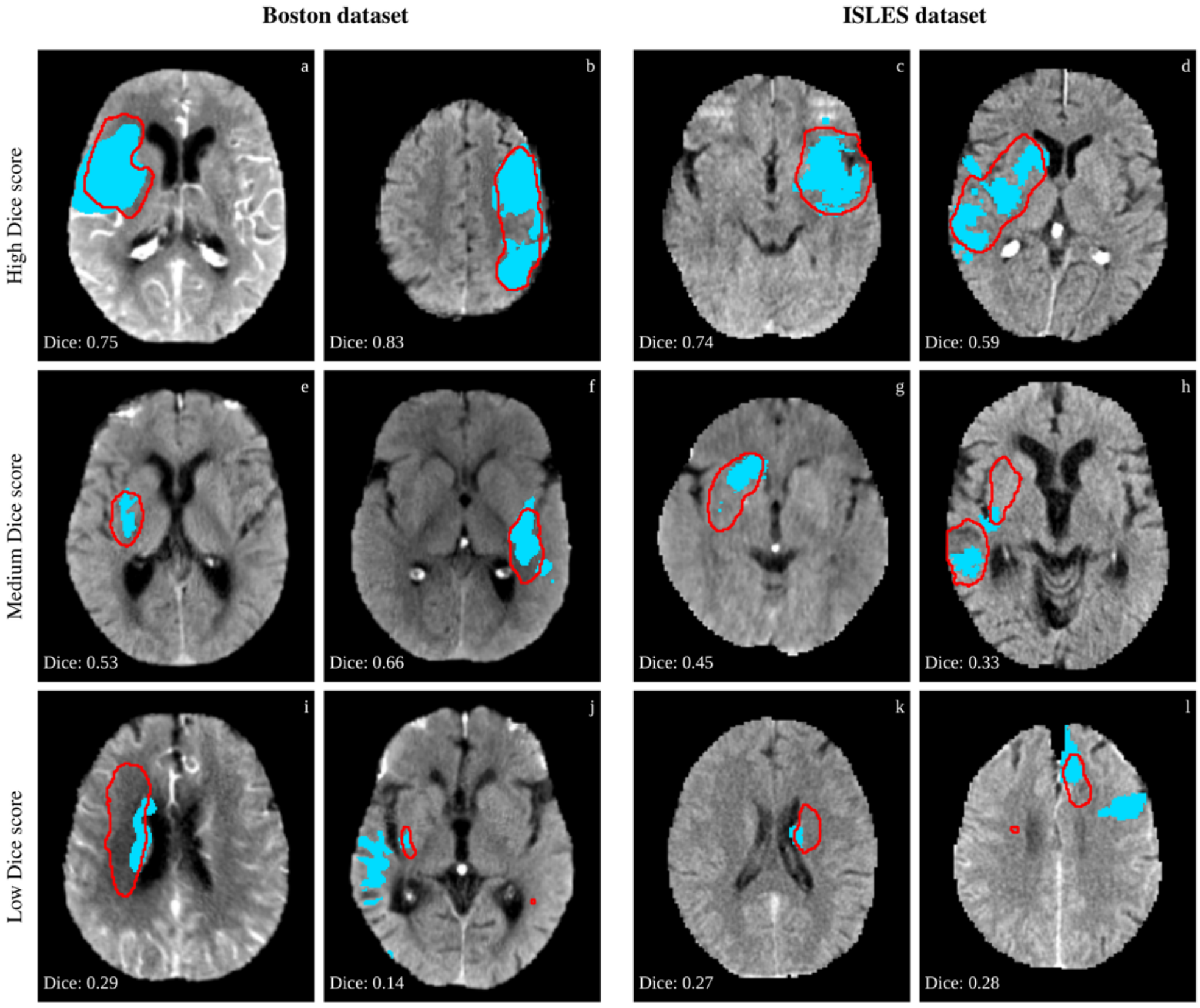

**Figure 4:** Twelve representative cases from the external test data are shown, including six Boston lesion segmentations and six ISLES final-infarct predictions, covering a range of performance levels (high, medium, and low Dice scores). For each case, the NCCT image is displayed with the reference lesion mask shown in cyan and the predicted segmentation outlined in red. Dice similarity coefficients are reported in the bottom left corner, and case identifiers are shown in the top right corner of each image.

### 3.3 Impact of Model Configurations

Internally (Table 4), mean Dice increased from 0.53 (conventional nnU-Net) to 0.57 (full configuration), particularly in AISD (0.47 to 0.52). Mirrored input alone did not change internal mean Dice but increased median Dice on the Boston dataset before fine-tuning from 0.23 to 0.41. Adding nnFoundation initialization raised it further to 0.47. After fine-tuning, median Dice increased stepwise from 0.40 to 0.56, with comparison modules and support masks contributing gains not observed before fine-tuning (0.45). The full configuration was used for all other analyses.

**Table 4:** Comparison of different pipeline configurations. Internal cross-validation performance (UKE + AISD, n = 595) and Boston test set performance (n = 46) before and after target-domain fine-tuning (FT) are shown for configurations stepwise adding mirrored CT input, nnFoundation initialization, comparison modules, and support masks. Boston results were obtained with models trained on all internal cases. Training schedules differed between configurations (see Methods). Metrics include Dice coefficient (internal: mean (SD); Boston: median [IQR]) and lesion detection rate (LDR). Bold values indicate the best performance across configurations.

| **Configuration** | **Internal cross-validation (n = 595)** | | | | **External test set: Boston (n = 46)** | |
|---|---|---|---|---|---|---|
| | **UKE Dice (mean (SD))** | **AISD Dice (mean (SD))** | **Overall Dice (mean (SD))** | **LDR (%)** | **Dice before FT (median [IQR])** | **Dice after FT (median [IQR])** |
| Native nnUNet | 0.65 (0.21) | 0.47 (0.30) | 0.53 (0.29) | 92.9 | 0.23 [0.00–0.51] | 0.40 [0.05–0.58] |
| + Mirrored CT | 0.66 (0.21) | 0.47 (0.30) | 0.53 (0.29) | 90.4 | 0.41 [0.14–0.60] | 0.45 [0.19–0.62] |
| + nnFoundation | 0.65 (0.23) | 0.50 (0.29) | 0.55 (0.28) | 92.6 | **0.47 [0.11–0.67]** | 0.53 [0.25–0.69] |
| + Comparison modules | 0.66 (0.21) | 0.51 (0.29) | 0.56 (0.27) | 93.9 | 0.45 [0.04–0.65] | 0.55 [0.35–0.68] |
| + Support masks | **0.67 (0.20)** | **0.52 (0.28)** | **0.57 (0.27)** | **94.3** | 0.45 [0.15–0.64] | **0.56 [0.44–0.70]** |

### 3.4 Agreement of Automated and Reference NWU Values

Fig. 5 shows automated versus reference NWU for both datasets. In UKE, the MAE was 0.98 percentage points (SD 1.45), with a Lin's CCC of 0.83 (95% CI 0.68–0.91). Bland–Altman analysis showed a mean bias of −0.09 and 95% limits of agreement from −3.52 to 3.33 percentage points. The corresponding Boston results were an MAE of 1.24 percentage points (SD 1.08), a CCC of 0.93 (95% CI 0.86–0.96), and a mean bias of −0.19 with limits of agreement from −3.42 to 3.04 percentage points. Absolute NWU error was not significantly associated with reference lesion volume in the Boston dataset (Spearman's ρ = −0.23, 95% bootstrap CI −0.53 to 0.09; $p$ = 0.12).

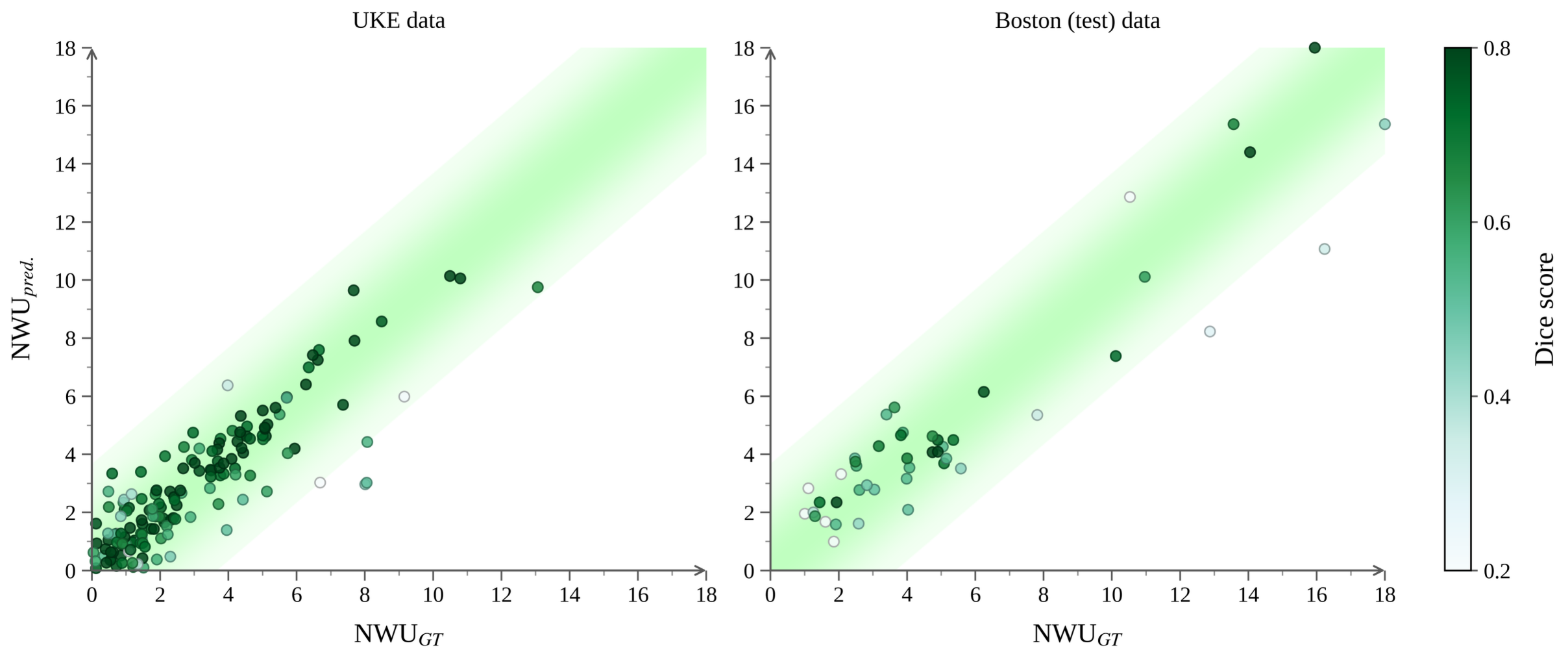

**Figure 5:** Reference NWU values (x-axis) compared to NWU values derived from the predicted lesion masks (y-axis) for the UKE dataset and the Boston held-out test set. Each point represents one case and is color-coded according to the corresponding Dice similarity coefficient between predicted and reference segmentations (darker green indicates higher Dice scores). The shaded diagonal region indicates an error margin of ±2 percentage points.

## 4. Discussion

We developed and externally validated a DL framework for ischemic stroke segmentation using exclusively NCCT imaging. By combining foundation-model initialization with explicit local comparisons of native and mirrored hemispheres, the framework utilizes anatomical domain knowledge while selective fine-tuning adapts the model to heterogeneous external sites. Importantly, relying solely on NCCT preserves the clinical practicality of a widely available first-line imaging modality, even though segmentation performance varied across domains and evaluation endpoints.

To evaluate cross-site generalizability under real-world acquisition shifts, we first benchmarked the framework on the external Boston cohort. On an identical held-out test set, our model achieved a median Dice score of 0.56, outperforming the 0.47 reported by Sentker et al. [9] and improving substantially over our own pre-adaptation baseline of 0.45. Component-wise ablation revealed distinct roles for each architectural choice: supplying a mirrored image copy yielded no internal gain but boosted out-of-domain performance on the Boston dataset prior to fine-tuning, confirming its value as a domain-invariant intra-scan reference. Foundation-model initialization added further baseline robustness, whereas the local comparison modules and support masks contributed modest internal gains (primarily in AISD) and improved performance on the Boston dataset only after fine-tuning, thus indicating that finer-grained feature matching requires brief target-domain calibration.

Beyond cross-institutional domain shift, clinical deployment also requires assessing whether acute NCCT features can predict downstream tissue outcome. We therefore evaluated the pipeline on the ISLES 2024 dataset, which shifts the task to a distinct longitudinal endpoint, that is, predicting final infarct extent 2–9 days post-intervention in successfully reperfused patients. The resulting mean Dice of 0.27 (SD 0.26) reflects the compounded effects of domain shift, subtle early NCCT visibility, and post-treatment tissue evolution, including potential procedure-related lesions [15, 16]. This aligns closely with the two top-ranked ISLES prediction models, which achieved mean Dice scores of 0.29 (SD 0.21) and 0.26 (SD 0.25) on the hidden challenge test set despite utilizing multimodal CT and perfusion maps [16, 24] – though differences in case composition and input modalities preclude direct comparison [15]. Nevertheless, a mean Dice of 0.27 remains insufficient for dependable patient-level final-infarct prediction. Dice was strongly associated with final-infarct volume, but its contribution to the Boston–ISLES performance difference cannot be isolated from reference timing and adaptation-set size. Minor boundary deviations or missed foci disproportionately affect overlap-based metrics in small lesions, while qualitative examples suggested additional challenges from multifocality. Dice should therefore be interpreted in the context of lesion visibility, size, and endpoint.

To contextualize our internal segmentation performance, we also compared our results against prior benchmarks on the AISD dataset. On the identical test split of Liang et al., our model achieved a mean Dice of 0.66, closely matching the 0.67 reported by Sun et al. and outperforming the original baseline of 0.58 [14, 25]. However, direct comparison is limited because our training set additionally incorporated UKE data. Notably, Sun et al. reported a drop to 0.49 on an external cohort [25], indicating performance loss under domain shift.

Beyond segmentation accuracy, clinical utility depends on whether automated masks support accurate downstream biomarkers such as NWU. NWU quantifies ischemic tissue water accumulation and predicts malignant infarction and infarct expansion, with values exceeding 12.7% carrying high prognostic relevance [6, 26]. Because NWU depends on mean attenuation across lesion and contralateral masks rather than exact boundary alignment, modest boundary deviations affect Dice far more than downstream quantification. Accordingly, automated NWU agreed closely with reference measurements across datasets, achieving a CCC of 0.93 and an MAE of 1.24 percentage points on the external Boston cohort – lower than the 1.48 percentage points previously reported on the same test set [9]. Moreover, absolute NWU error showed no significant association with lesion volume in the Boston dataset. Deriving both lesion segmentation and quantitative NWU from baseline NCCT alone therefore offers a streamlined alternative when advanced perfusion imaging is unavailable.

Several limitations should be considered when interpreting these findings. First, supervised fine-tuning required annotated target-domain subsets (11 Boston and 75 ISLES cases), creating a logistical bottleneck for multicenter deployment that unsupervised or self-supervised domain adaptation could mitigate. Second, NCCT-only segmentation remains limited for small, multifocal, or hyperacute infarcts where parenchymal hypoattenuation is subtle or not yet established; future integration of clinical metadata, such as time from symptom onset, or serial imaging may improve sensitivity in these settings. In addition, resampling heterogeneous scans to 1.0 × 1.0 × 3.0 mm cannot restore information absent from thick-slice acquisitions and may introduce partial-volume averaging, although symmetric preprocessing of lesion and contralateral masks likely minimizes systematic bias in NWU. Finally, reference standards varied across cohorts – ranging from pretreatment CTP-derived perfusion deficits (UKE) and pre-decision DWI (Boston) to postinterventional final infarcts (ISLES) – which are not biologically interchangeable and may introduce cross-modal registration discrepancies independent of model error.

In summary, domain-specific adaptation enabled robust NCCT-only infarct segmentation across heterogeneous external cohorts, although performance varied across clinical endpoints. Importantly, automated lesion masks supported accurate, low-error NWU quantification despite moderate spatial overlap, warranting prospective studies to establish its prognostic and clinical utility.

**Availability of data and materials**

The AISD and ISLES 2024 datasets are publicly available as described in the cited dataset publications. The UKE and Boston datasets are not publicly available due to patient confidentiality restrictions. Source code will be released at github.com/IPMI-ICNS-UKE/ncct-stroke-seg after acceptance.

## References

1. Campbell BCV, Khatri P (2020) Stroke. The Lancet 396:129–142

2. Czap AL, Sheth SA (2021) Overview of Imaging Modalities in Stroke. Neurology 97:42–51

3. Demeestere J, Verhaaren BFJ, Christensen S et al (2025) Underestimation of Follow-Up Infarct Volume by Acute CT Perfusion Imaging. Neurology 104:e213439

4. Potreck A, Weyland CS, Seker F et al (2022) Accuracy and Prognostic Role of NCCT-ASPECTS Depend on Time from Acute Stroke Symptom-onset for both Human and Machine-learning Based Evaluation. Clin Neuroradiol 32:133–140

5. Neumann AB, Jonsdottir KY, Mouridsen K et al (2009) Interrater Agreement for Final Infarct MRI Lesion Delineation. Stroke 40:3768–3771

6. Broocks G, Flottmann F, Scheibel A et al (2018) Quantitative Lesion Water Uptake in Acute Stroke Computed Tomography Is a Predictor of Malignant Infarction. Stroke 49:1906–1912

7. Ghozy S, Amoukhteh M, Hasanzadeh A et al (2024) Net water uptake as a predictive neuroimaging marker for acute ischemic stroke outcomes: a meta-analysis. Eur Radiol 34:5308–5316

8. Broocks G, Bendszus M, Simonsen C et al (2025) Net Water Uptake at CT Predicts the Treatment Effect of Thrombectomy for Low ASPECTS Stroke. Radiology 317:e250708

9. Sentker T, Nielsen M, Klapproth S et al (2025) Fully automated quantification of net water uptake in acute ischemic stroke using only non-contrast CT imaging. Eur Radiol 36:4976–4986

10. Matta S, Lamard M, Zhang P et al (2024) A systematic review of generalization research in medical image classification. Comput Biol Med 183:109256

11. Welland SH, Kim GHJ, Yadav A et al (2026) Characterizing the effects of noncontrast head CT reconstruction kernel and slice thickness parameters on the performance of an automated AI algorithm in the evaluation of ischemic stroke. J Med Imaging 13:e014503

12. AlBadawy EA, Saha A, Mazurowski MA (2018) Deep learning for segmentation of brain tumors: Impact of cross-institutional training and testing. Med Phys 45:1150–1158

13. Isensee F, Jaeger PF, Kohl SAA, Petersen J, Maier-Hein KH (2021) nnU-Net: a self-configuring method for deep learning-based biomedical image segmentation. Nat Methods 18:203–211

14. Liang K, Han K, Li X et al (2021) Symmetry-Enhanced Attention Network for Acute Ischemic Infarct Segmentation with Non-contrast CT Images. In: De Bruijne M, Cattin PC, Cotin S, et al (eds) Medical Image Computing and Computer Assisted Intervention – MICCAI 2021. Springer International Publishing, Cham, pp 432–441

15. Riedel EO, De La Rosa E, Baran TA et al (2026) The Ischemic Stroke Lesion Segmentation Challenge (ISLES)'24 Dataset: A Multimodal Stroke Imaging Dataset with Hyperacute CT, Acute Postinterventional MRI, and 3-month Clinical Outcomes. Radiol Artif Intell 8:e250603

16. de la Rosa E, Su R, Reyes M et al (2024) ISLES'24: Final Infarct Prediction with Multimodal Imaging and Clinical Data. Where Do We Stand? Preprint at https://doi.org/10.48550/ARXIV.2408.10966

17. Muschelli J (2020) A Publicly Available, High Resolution, Unbiased CT Brain Template. In: Lesot M-J, Vieira S, Reformat MZ, et al (eds) Information Processing and Management of Uncertainty in Knowledge-Based Systems. Springer International Publishing, Cham, pp 358–366

18. Wasserthal J, Breit H-C, Meyer MT et al (2023) TotalSegmentator: Robust Segmentation of 104 Anatomic Structures in CT Images. Radiol Artif Intell 5:e230024

19. Harsy CU, Wald T, Gotkowski K et al (2026) nnFoundation: 3D Foundation Models for Radiology. Preprint at https://doi.org/10.48550/arXiv.2609.26924

20. Shah M, Sundaresan V (2026) AsymFeX: A Symmetry-Driven Framework for Ischemic Stroke Segmentation Across Imaging Modalities and Stroke Stages. Preprint at https://doi.org/10.48550/arXiv.2608.19769

21. Lee Y, Chen AS, Tajwar F et al (2023) Surgical Fine-Tuning Improves Adaptation to Distribution Shifts. In: International Conference on Learning Representations (ICLR 2023)

22. Karani N, Chaitanya K, Baumgartner C, Konukoglu E (2018) A Lifelong Learning Approach to Brain MR Segmentation Across Scanners and Protocols. In: Frangi AF, Schnabel JA, Davatzikos C, Alberola-López C, Fichtinger G (eds) Medical Image Computing and Computer Assisted Intervention – MICCAI 2018. Springer International Publishing, Cham, pp 476–484

23. Padilla R, Netto SL, Da Silva EAB (2020) A Survey on Performance Metrics for Object-Detection Algorithms. In: 2020 International Conference on Systems, Signals and Image Processing (IWSSIP). IEEE, Niterói, Brazil, pp 237–242

24. Ren T, Rivera JEH, Oswal H et al (2025) How We Won the ISLES'24 Challenge by Preprocessing. Preprint at https://doi.org/10.48550/ARXIV.2505.18424

25. Sun J, Ju G-L, Qu Y-H et al (2026) Deep Learning for Segmenting Ischemic Stroke Infarction in Non-contrast CT Scans by Utilizing Asymmetry. Clin Neuroradiol 36:129–142

26. Marcus A, Mair G, Chen L et al (2024) Deep learning biomarker of chronometric and biological ischemic stroke lesion age from unenhanced CT. Npj Digit Med 7:338